\documentclass[10pt,twocolumn,letterpaper]{article}

\usepackage[pagenumbers]{iccv} 

\usepackage{graphicx}
\usepackage{caption}
\usepackage{amsmath, amsfonts, amssymb}
\usepackage{colortbl}
\usepackage{pifont}
\usepackage{float}
\definecolor{myorange}{RGB}{255,165,0} 
\definecolor{myyellow}{RGB}{255,255,0} 

\definecolor{iccvblue}{rgb}{0.21,0.49,0.74}
\usepackage[pagebackref,breaklinks,colorlinks,allcolors=iccvblue]{hyperref}

\def\paperID{5576} 
\def\confName{ICCV}
\def\confYear{2025}

\title{MF-VITON: High-Fidelity Mask-Free Virtual Try-On with Minimal Input}

\author{
Zhenchen Wan$^1$ \quad Yanwu Xu \quad Dongting Hu$^1$ \quad Weilun Cheng$^1$ \quad Tianxi Chen$^1$ \\ 
Zhaoqing Wang$^2$ \quad Feng Liu$^1$ \quad Tongliang Liu$^2$ \quad Mingming Gong$^{1,3}$ \\
$^1$University of Melbourne, Melbourne, Australia \\
$^2$The University of Sydney, Sydney, Australia \\
$^3$Mohamed bin Zayed University of Artificial Intelligence, Abu Dhabi, UAE \\
{\tt\small \{zhenchenw, dongting, weilunjc, tianxic, fengliu.ml, mingming.gong\}@unimelb.edu.au} \\
{\tt\small zwan6779@uni.sydney.edu.au, tongliang.liu@sydney.edu.au}
}

\begin{document}

\twocolumn[{%
\renewcommand\twocolumn[1][]{#1}%
\maketitle
\begin{center}
    \centering
    \includegraphics[width=\textwidth]{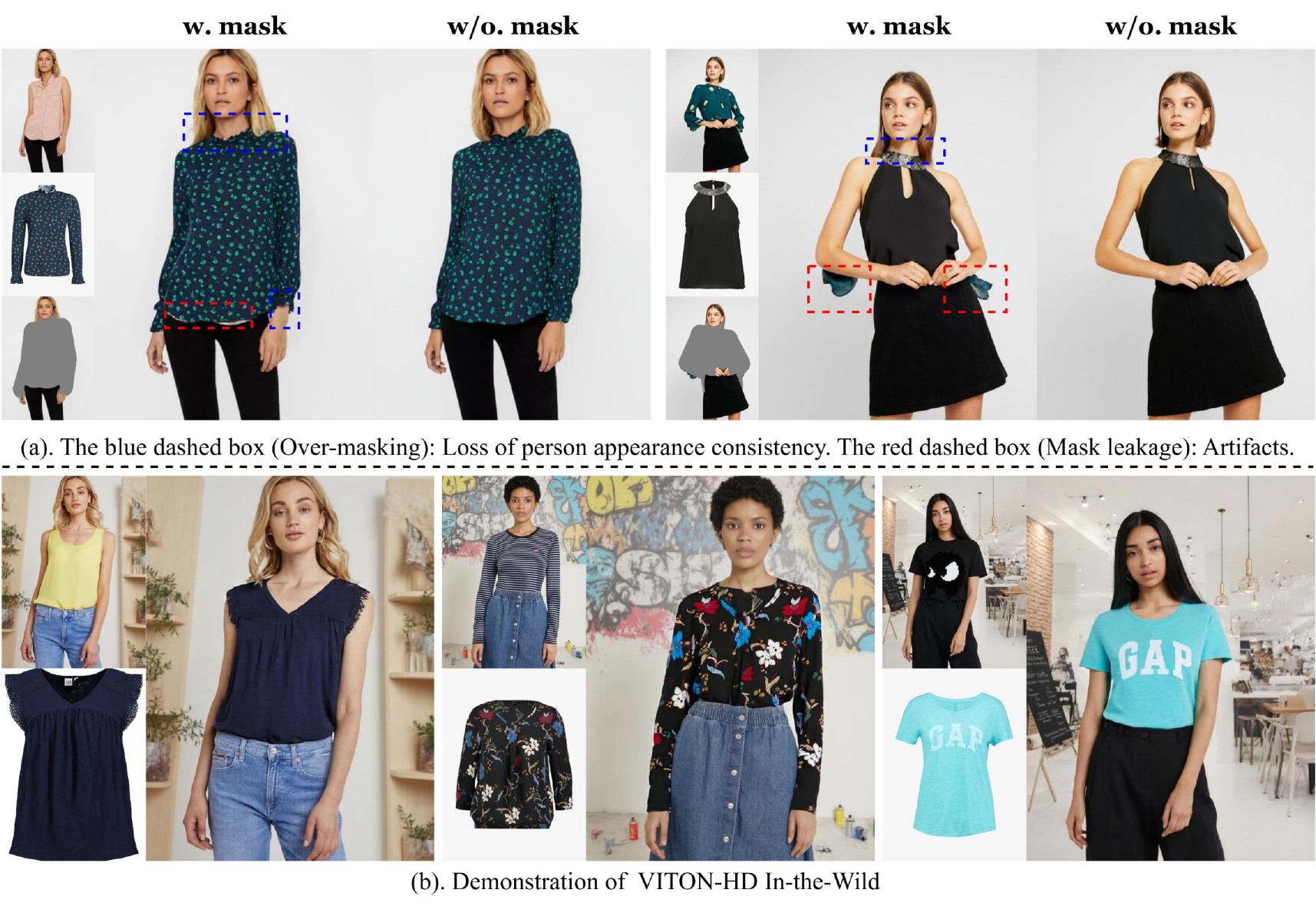}
    \captionsetup{type=figure, skip=-10pt}
    \captionof{figure}{We propose a Mask-Free Virtual Try-On framework that achieves SOTA visual quality by eliminating artifacts caused by inaccurate masks: (a) Eliminate interference from inaccurate masks: Inaccurate masks cause over-masking, leading to unnatural regeneration of hair or hands, and mask leakage, resulting in artifacts like remnants of the old clothing. (b) Demonstration of VITON-HD In-the-Wild.}
    \label{fig:cover_image}
    \vspace{-0.5em}
\end{center}%
}]

\begin{abstract}

\vspace{-1.0em}

Recent advancements in Virtual Try-On (VITON) have significantly improved image realism and garment detail preservation, driven by powerful text-to-image (T2I) diffusion models. However, existing methods often rely on user-provided masks, introducing complexity and performance degradation due to imperfect inputs, as shown in Fig.~\ref{fig:cover_image}(a). To address this, we propose a Mask-Free VITON (MF-VITON) framework that achieves realistic VITON using only a single person image and a target garment, eliminating the requirement for auxiliary masks. Our approach introduces a novel two-stage pipeline: (1) We leverage existing Mask-based VITON models to synthesize a high-quality dataset. This dataset contains diverse, realistic pairs of person images and corresponding garments, augmented with varied backgrounds to mimic real-world scenarios. (2) The pre-trained Mask-based model is fine-tuned on the generated dataset, enabling garment transfer without mask dependencies. This stage simplifies the input requirements while preserving garment texture and shape fidelity. Our framework achieves state-of-the-art (SOTA) performance regarding garment transfer accuracy and visual realism. Notably, the proposed Mask-Free model significantly outperforms existing Mask-based approaches, setting a new benchmark and demonstrating a substantial lead over previous approaches.
\end{abstract}    
\section{Introduction}
\label{sec:intro}

Virtual Try-On (VITON) technology has transformed consumer engagement with fashion by enabling them to visualize how garments would look on them without the need for physical fitting. Traditionally, VITON systems rely on input images of an individual, a selected garment, and a precise mask indicating the region to be inpainted, generating realistic images of the individual wearing the chosen attire \cite{choi_viton-hd_2021, wan_improving_2024, choi_improving_2024, wan_ted-viton_2024}. However, despite its transformative potential, VITON faces several critical challenges that hinder its widespread adoption and effectiveness. These challenges stem from the inherent complexity of accurately aligning garments with diverse body shapes and poses, preserving fine-grained details such as textures and logos, and generating high-quality, artifact-free images \cite{choi_improving_2024, wan_ted-viton_2024, morelli_ladi-vton_2023}. Additionally, the reliance on precise user-provided masks and the scarcity of high-quality, diverse datasets further complicate the development of robust and generalizable VITON systems. Addressing these challenges is crucial for advancing VITON technology and unlocking its full potential in real-world applications.

The primary challenges in VITON can be categorized into three main areas: (1) Precision in Masking, where the system must generate highly accurate masks to delineate the regions where the garment will be placed \cite{zhu_mm_2024, li_unihuman_2023, wan_improving_2024, choi_improving_2024, wan_ted-viton_2024}; (2) Predictive Masking, which involves the ability to predict and adapt the mask to the specific areas of the body where the garment will be worn \cite{gou_taming_2023, morelli_ladi-vton_2023, ning_picture_2024, kim_stableviton_2023}; and (3) Data Generation, which relates to the difficulty in creating high-quality, diverse datasets that can effectively train VITON models \cite{dong_fw-gan_2019, han_viton_2018, bai_single_2022,  choi_viton-hd_2021, minar_cp-vton_2020}. These challenges are compounded by the need for models to generalize well across different body types, poses, and garment styles.

To address these challenges, we propose a novel framework that leverages recent advancements in VITON technology while introducing innovative solutions to overcome the aforementioned limitations. Our contributions are as follows:

\begin{itemize}
\item \textbf{Reduced Conditional Input Requirements:} Our framework significantly reduces the reliance on predefined masks and other conditional inputs, simplifying the VITON process and improving usability.
\item \textbf{Dataset Creation Methodology and Dataset Provision:} We propose an innovative dataset generation pipeline that utilizes a pre-trained Mask-based VITON model (e.g., IDM-VTON \cite{choi_improving_2024} and TED-VITON \cite{wan_ted-viton_2024}) to produce high-quality garment-person images. To further enhance background diversity while preserving garment-person consistency, we incorporate text-guided inpainting models \cite{flux2024}, enabling a more robust and adaptable training dataset.
\item \textbf{Output-for-Input (OFI) Strategy for Robust Training:} We propose an OFI training strategy that leverages the outputs of Mask-based models, even when generated with inaccurate masks, as inputs for training Mask-Free models. This approach effectively introduces noise during training, enhancing the robustness of Mask-Free models and reducing their reliance on precise masks. Our experiments demonstrate that OFI not only mitigates artifacts caused by mask inaccuracies but also improves generalization, sometimes achieving better results than those trained on clean data.
\item \textbf{Plug-and-Play Integration with ReferenceNet:} We introduce a plug-and-play strategy that can be seamlessly integrated into existing VITON models that utilize ReferenceNet \cite{choi_improving_2024, wan_ted-viton_2024}. With minimal fine-tuning, this enhancement significantly improves realism and garment alignment in diverse real-world scenarios.
\end{itemize}

By addressing these fundamental challenges, our framework pushes the boundaries of VITON technology, making it more accurate, efficient, and adaptable for practical applications. Extensive evaluations on VITON-HD \cite{choi_viton-hd_2021}, DressCode \cite{morelli_dress_2022}, and VITON-HD In-the-Wild demonstrate substantial improvements, achieving up to a 34.7\% reduction in FID and KID scores compared to prior Mask-based methods. Moreover, as a Plug-and-Play solution, our approach is compatible with a wide range of existing VITON models, significantly reducing their reliance on masks while preserving or even enhancing their original performance. The following sections provide an in-depth discussion of our framework, detailing its methodologies, technical innovations, and experimental validation.
\section{Related Works}
\label{sec:related}

\noindent\textbf{Pose-Guided Person Image Synthesis (PPIS).} VITON technology has its roots in PPIS, which focuses on generating person images conditioned on specific body poses. Early PPIS approaches aimed to create visually convincing images of individuals in various postures, laying the groundwork for VITON. Pioneering works in this domain \cite{ma_pose_2017, liu_liquid_2019, zhu_progressive_2019, zhou_cross_2022, fruhstuck_insetgan_2022, albahar_pose_2021, li_pona_2020, men_controllable_2020} focused on aligning human poses with target clothing images, addressing key challenges in pose transfer and adaptation to individual body shapes.

\noindent\textbf{Mask-based VITON.} Mask-based VITON approaches, whether GAN-based or diffusion-based, depend on precise masks to delineate regions for garment placement and inpainting. GAN-based methods \cite{liu_liquid_2019, fruhstuck_insetgan_2022, kips_ca-gan_2020, dong_fw-gan_2019, honda_viton-gan_2019, raffiee_garmentgan_2021, pecenakova_fitgan_2022, albahar_pose_2021, men_controllable_2020, xie_gp-vton_2023, lee_high-resolution_2022, shim_towards_2024} typically involve deforming garments to match the target person’s body shape and fusing them with the person’s image. However, these methods often struggle with generalization in complex backgrounds and varied poses. Diffusion-based VITON methods \cite{cui_street_2024, bhatnagar_multi-garment_2019, zhu_tryondiffusion_2023, ning_picture_2024, li_unihuman_2023, morelli_ladi-vton_2023, gou_taming_2023, kim_stableviton_2023, wan_improving_2024, choi_improving_2024, wan_ted-viton_2024} have emerged as a promising alternative, leveraging noise-reversal processes to enhance garment fidelity and detail preservation. For instance, StableVITON \cite{kim_stableviton_2023} and IDM-VTON \cite{choi_improving_2024} utilize ControlNet and IP-Adapter, respectively, to improve garment-body alignment and fine-grained detail preservation. Despite these advancements, preserving intricate elements like logos and textures under diverse poses and lighting conditions remains challenging.

\noindent\textbf{Localized Control and Data Generation.} Recent research has explored localized control mechanisms, such as SEED-Edit \cite{shi_seededit_2024}, which enable precise manipulation of specific regions in generated images. These methods have inspired our approach to dataset generation, where we leverage generative models to create synthetic data for training. For example, BooW-VTON \cite{zhang_boow-vton_2024} introduces a Mask-Free training paradigm, using pseudo-data augmentation to enhance model performance in complex scenarios without requiring precise masks. These innovations highlight the potential of generative models in creating high-quality datasets for training robust VITON systems.

By integrating these advances, our work aims to address the limitations of existing VITON methods, particularly in mask precision, garment detail preservation, and dataset generation, while leveraging localized control mechanisms to enhance model performance and adaptability.
\section{Preliminary}
\label{sec:Preliminary}

\textbf{Stable Diffusion} \cite{rombach_high-resolution_2022} is a latent diffusion model that learns a denoising process in a lower-dimensional latent space, improving efficiency while maintaining high fidelity. It consists of a latent encoder-decoder, a denoiser, and a noise scheduler, facilitating realistic image generation from textual or visual inputs. The model is trained using conditional flow matching (CFM) loss \cite{esser_scaling_2024}:  
\begin{equation} 
    \mathcal{L}_{\text{CFM}} = \mathbb{E}_{t, z, \epsilon} \Big[ \lambda(t) \cdot \big\| v_\theta(z_t, t) - \nabla_z \log p_t(z_t | X_{\text{target}}) \big\|_2^2 \Big],
\end{equation}  
where $z_t$ is the latent variable at timestep $t$, and $v_\theta$ estimates the velocity field. The weighting function $\lambda(t)$ stabilizes training. Unlike standard diffusion models that perform noise regression, CFM learns data transport trajectories, achieving 2.1$\times$ faster convergence while ensuring Lipschitz continuity. The final image is reconstructed via Euler-Maruyama integration: $X_{\text{out}} = \int_{t=1}^0 v_\theta(z_t, t) \, dt$.

\section{Method}
\label{sec:method}

\begin{figure*}[ht]
    \centering
    \includegraphics[width=1\textwidth]{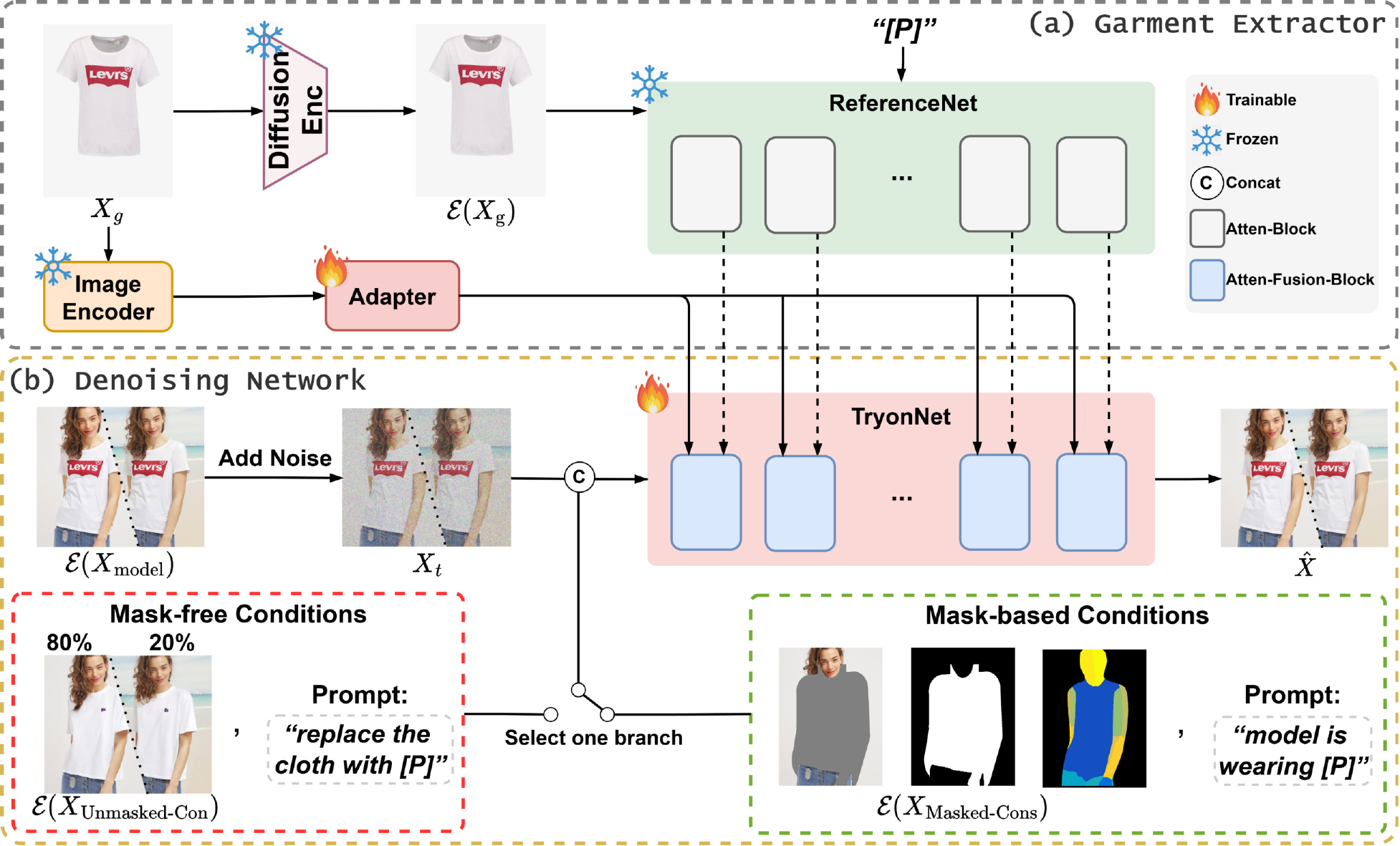}
    \caption{\textbf{Overview of MF-VITON:} We propose a \textbf{Mask-based \& Mask-Free VITON pipeline} that enables seamless adaptation from Mask-based to MF-VITON. The pipeline comprises two branches: (a) the \textbf{Garment Extractor}, which leverages \textbf{ReferenceNet} to encode fine-grained garment features $\mathcal{E}(x_{\text{g}})$ and employs an \textbf{Adapter} \cite{ye_ip-adapter_2023} to extract high-level semantics from garment images $X_g$ using a pretrained image encoder; and (b) the \textbf{Denoising Network}, which utilizes \textbf{TryonNet} as the primary denoising branch to process concatenated inputs of noised latent $X_t$ and selectively integrates either \textbf{Mask-based conditions} ($\mathcal{E}(X_{\text{Masked-Con}})$, Mask-based text prompt) or \textbf{Mask-Free conditions} ($\mathcal{E}(X_{\text{Unmasked-Con}})$, Mask-Free text prompt). }
    \label{fig:MF-VITON}
    \vspace{-1.5em} 
\end{figure*}

\noindent\textbf{Model Overview.} Fig.~\ref{fig:MF-VITON} illustrates the Mask-based \& MF-VITON pipeline, a two-stage framework designed to achieve high-quality MF-VITON with minimal input requirements. The framework is compatible with SOTA ReferenceNet-based VITON models such as IDM-VTON \cite{choi_improving_2024} and TED-VITON \cite{wan_ted-viton_2024}, both of which consist of three core components: ReferenceNet, Adapter, and TryonNet. The pipeline operates in two stages: Stage I: Mask-based VITON Training \& Dataset Generation, where a dataset is constructed by ensuring that the model identity, pose, and other conditions remain unchanged, while only the clothing varies. Stage II: MF-VITON Training, where the model is fine-tuned using the Mask-Free dataset to generalize well with minimal input constraints. 

\subsection{Mask-based \& MF-VITON pipeline} 

The pipeline remains unchanged across both stages, with the key difference lying in the nature of the input data. Following TED-VITON~\cite{wan_ted-viton_2024}, a large language model is employed to generate a descriptive representation of the given garment $X_g$, enhancing the model’s understanding of the garment. 

\noindent\textbf{ReferenceNet}, as shown in Fig.~\ref{fig:MF-VITON}(a), is designed to extract fine-grained garment features, including textures, patterns, fabric structures, logos, and other subtle details critical for realistic VITON results, which processes the garment image $ X_g $ through a frozen pre-trained VAE encoder to obtain its latent representation $ \mathcal{E}(X_g) $, which is combined with the conditioned text prompt $ \tau_\theta(P) $ generated by the text encoder. These representations are further refined through multiple network layers to preserve intricate garment characteristics.

Formally, the ReferenceNet branch processes the garment image $ X_g $ and the conditioned text prompt $ P $ as follows:
\begin{equation}
    F_{\text{reference}}^i = \text{ReferenceNet}^i(\mathcal{E}(X_g), \tau_\theta(P)),
\end{equation}
where $ F_{\text{reference}}^i $ denotes the fine-grained garment features extracted at the $ i $-th layer of ReferenceNet.

\noindent\textbf{Adapter} \cite{ye_ip-adapter_2023}, illustrated in Fig.~\ref{fig:MF-VITON}(a), enhances model generalization by mitigating sensitivity to body poses, garment deformations, and environmental factors. It achieves this by focusing on low-frequency garment attributes, ensuring robustness across diverse conditions. Unlike ReferenceNet, the Adapter extracts semantic garment information, including structure, style, and material properties, using a frozen pre-trained image encoder. The extracted high-order semantics, denoted as $ H_{\text{semantic}} $, provide a broader contextual understanding of clothing while enhancing adaptability.  

To effectively fuse these features, the Adapter employs a decoupled cross-attention mechanism, separately processing joint and image embeddings. Given a query matrix $ \mathbf{Q} \in \mathbb{R}^{N \times d} $, with key-value pairs $ (\mathbf{K}_j, \mathbf{V}_j) $ for joint embeddings and $ (\mathbf{K}_i, \mathbf{V}_i) $ for image embeddings, the output is computed as:
\begin{equation}
    \mathbf{Z}_{\text{new}} = \text{Attention}(\mathbf{Q}, \mathbf{K}_j, \mathbf{V}_j) + \lambda \cdot \text{Attention}(\mathbf{Q}, \mathbf{K}_i, \mathbf{V}_i),
    \label{eq:GS-Adapter}
\end{equation}
where $ \lambda $ controls the relative contributions of image and joint features. This formulation enhances the model’s robustness in handling pose variations, garment structures, and environmental shifts, ensuring photorealistic try-on results.

\noindent\textbf{TryonNet}, as depicted in Fig.~\ref{fig:MF-VITON}(b), operates in a VAE-constrained latent space, processing two distinct input streams: the Mask-based Context ($ \zeta $) for precise garment replacement and the Mask-Free Context ($ \xi $) for enhanced generalization. The Mask-based Context is a composite input defined as $ \zeta = [\mathcal{E}(X_{\text{model}}); m; \mathcal{E}(X_{\text{mask}}); \mathcal{E}(X_{\text{pose}})] $, which includes: (1) the latent representation of the person for structural guidance, (2) a dynamically resized garment mask, (3) the masked person features for reconstruction, and (4) pose embeddings for alignment. The Mask-Free Context, defined as $\xi = [\mathcal{E}(X_{\text{model}}); \mathcal{E}(X_{\text{model'}})]$, enhances the model’s robustness for In-the-Wild scenarios. Here, $X_{\text{model}}$ consists of 80\% instances where the same model appears with different clothing and 20\% instances where both the clothing and background are altered, providing a balanced augmentation strategy to improve generalization.

\noindent\textbf{Fusion Mechanism} combines reference features $F_{\text{reference}}^i$ extracted from ReferenceNet with try-on features $F_{\text{tryon}}^i$ through the attention process. Text embeddings $\tau_\theta(P)$ are integrated into the attention mechanism via concatenated query-key-value projections. This is further enriched by high-order semantic features $H_{\text{semantic}}$ from the Adapter, resulting in an unified feature representation. The final virtual try-on output is generated through:  
\begin{equation}
    \hat{X} = \text{TryonNet}(\{\zeta, \xi\}, \tau_\theta(P), F_{\text{reference}}, H_{\text{semantic}}).
\end{equation}

\begin{figure*}[ht]
    \centering
    \includegraphics[width=1\textwidth]{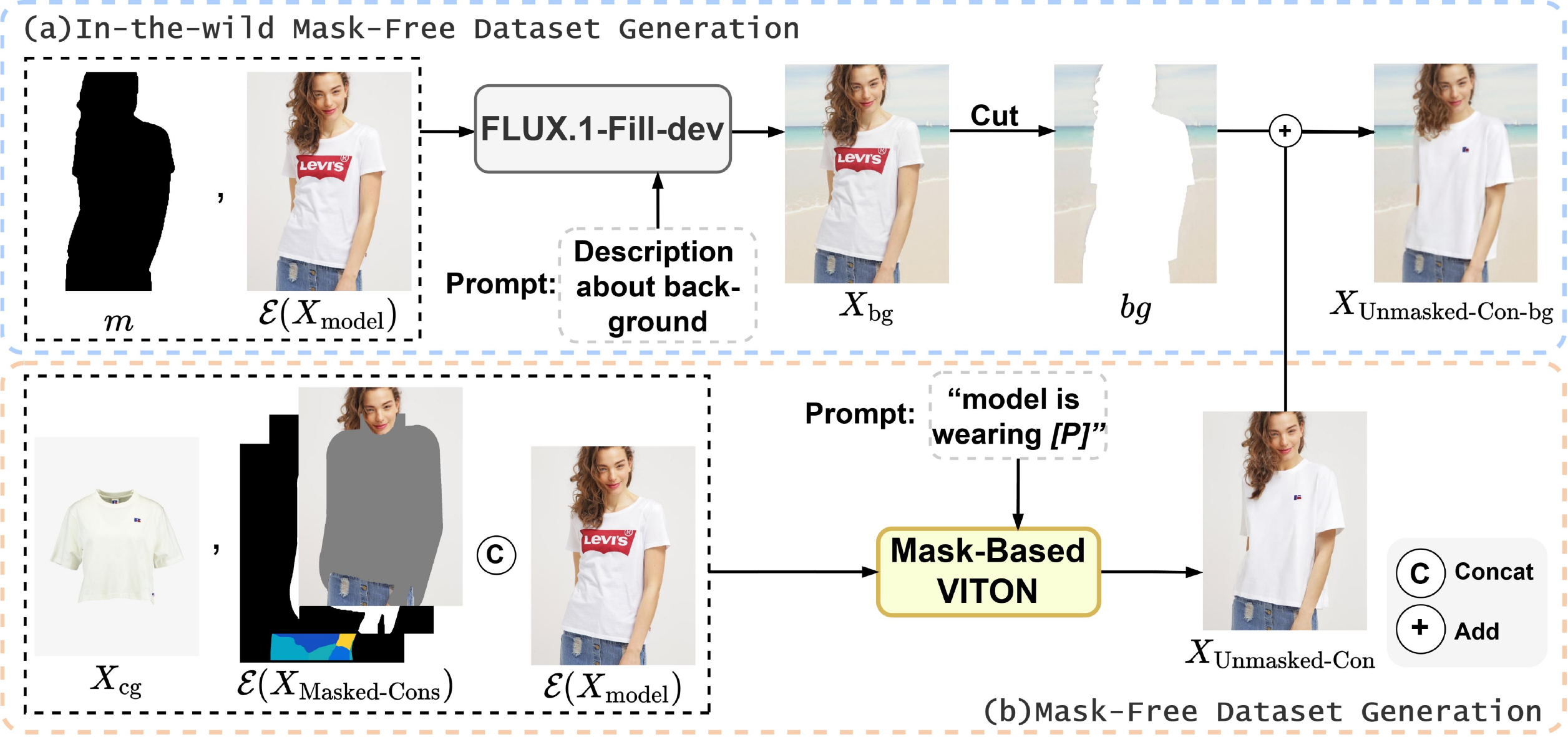}
    \caption{\textbf{Overview of MF-VITON Dataset Generation:} (a) \textbf{In-the-Wild Mask-Free Dataset Generation}: Uses FLUX.1-Fill-dev \cite{flux2024} to generate realistic background-filled model images $X_{\text{bg}}$, which are then composited with the Mask-based background $bg$ to create Mask-Free dataset samples $X_{\text{Unmasked-Con-bg}}$. (b) \textbf{Mask-Free Dataset Generation}: Concatenates the noised latent encoding $\mathcal{E}(X_{\text{model}})$ with Mask-based conditions $\mathcal{E}(X_{\text{Masked-Cons}})$. The Mask-based VITON model then synthesizes garment-swapped images $X_{\text{Unmasked-Con}}$.}
    \label{fig:dataset_generation}
\end{figure*}

\subsection{Mask-Free Dataset Preparation}

The primary challenge in developing MF-VITON lies in the incompleteness of the dataset. Specifically, if the conditional input to the model $\mathcal{E}(X_{\text{Unmasked-Con}})$ is identical to the output image $\hat{X}$, where both the input and output depict the model wearing the same garment, the VITON model will fail to learn how to change the target garment $X_g$. This is due to there exists a short path where the model can directly copy the input without learning the garment transformation. To address this, the dataset must consist of images where all conditions (e.g., pose, background, and model identity) remain consistent except for the garment.

To generate the Mask-Free conditional input $X_{\text{Unmasked-Con}}$, we leverage TryonNet by utilizing the Mask-based Context ($\zeta$) and the corresponding Mask-based prompt as input. This process synthesizes an image in which the target garment $X_{\text{cg}}$ is realistically integrated into the model while preserving other visual attributes. The target garment $X_{\text{cg}}$ is separately processed through ReferenceNet and Adapter to ensure accurate texture and structural preservation. Formally, this can be expressed as:
\begin{equation}
    X_{\text{Unmasked-Con}} = \text{TryonNet}(\zeta, \tau_\theta(m), F_{\text{reference}}, H_{\text{semantic}}),
\end{equation}
where $\tau_\theta(m)$ represents the encoded Mask-based prompt, $F_{\text{reference}}$ denotes the extracted fine-grained garment features from ReferenceNet, and $H_{\text{semantic}}$ encapsulates the structural and stylistic garment information captured by the Adapter.

In this setting, the conditional input $\mathcal{E}(X_{\text{Unmasked-Con}})$ provides information about the regions that should remain unchanged, while the target garment $X_g$ ensures that the model learns to synthesize the desired clothing transformation. This dataset construction approach effectively prevents the model from merely copying input information, thereby enforcing garment adaptation and transformation learning.

\subsection{In-the-Wild Dataset Augmentation}

Previous works have collected In-the-Wild datasets from the internet \cite{choi_improving_2024, cui_street_2024, zhang_boow-vton_2024}, which is challenging due to the difficulty of obtaining a large number of high-quality images. However, recent advancements in inpainting models have significantly improved performance, allowing them to preserve unmasked regions while generating high-quality content for masked areas. Leveraging this, we utilize the SOTA inpainting model, Flux.1-fill-dev \cite{flux2024}, to augment the VITON-HD \cite{choi_viton-hd_2021} dataset by filling in the blank backgrounds, thereby creating In-the-Wild images.

The augmentation process begins with background inpainting. Given the original model image $\mathcal{E}(X_{\text{model}})$, the background mask $m$, and a textual prompt describing the desired background, Flux.1-fill-dev \cite{flux2024} generates a new image $X_{\text{bg}}$ with a realistic background, as described by the equation:
\begin{equation}
    X_{\text{bg}} = \text{Flux.1-fill-dev}\left(\mathcal{E}(X_{\text{model}}), m, \text{prompt}\right).
    \label{eq:Flux-fill}
\end{equation}
Next, for Mask-Free training, we construct data pairs where the model wears different garments while preserving the same background. This is achieved by utilizing the background mask \(m\) and the in-painted image \(X_{\text{bg}}\) to extract the background \(bg\), which is then composited with the Mask-Free condition image \(X_{\text{Unmasked-Con}}\) to generate the Mask-Free condition image with background \(X_{\text{Unmasked-Con-bg}}\), ultimately forming a data pair \((X_{\text{Unmasked-Con-bg}}, X_{\text{bg}})\). This approach ensures that the model learns to focus on garment transformation while maintaining background consistency. By enriching the dataset with diverse In-the-Wild scenarios, this augmentation process enhances the model’s ability to generalize to real-world applications.

\subsection{Output-for-Input Training Strategy}

Despite significant advancements in Mask-based VITON models, they still struggle in extreme scenarios, such as rare body poses or inaccurate masks, leading to artifacts that fail to match real-world image quality. To address this, we introduce the OFI training strategy, which leverages the results generated by the Mask-based VITON model as training inputs for the Mask-Free model. By ensuring the model learns from synthesized yet realistic images, OFI effectively raises its performance ceiling. Since the inherent errors of the Mask-based model are integrated into the training input, while still providing all necessary conditions except for the garment, these errors do not significantly degrade performance.

Formally, given the Mask-based VITON output \(X_{\text{masked-out}}\) from Stage I, defined as \(X_{\text{masked-out}} = \text{TryonNet}(\zeta, \tau_\theta(P), F_{\text{reference}}, H_{\text{semantic}})\), we set the unmasked conditional input as \(X_{\text{Unmasked-Con}} = X_{\text{masked-out}}\). The final MF-VITON output is then generated as $\hat{X} = \text{TryonNet}(X_{\text{Unmasked-Con}}, \tau_\theta(P), F_{\text{reference}}, H_{\text{semantic}}).$ 
This strategy avoids the interference caused by low-quality outputs from Mask-based VITON, leading to more natural and artifact-free results that better align with real-world imagery.

\begin{figure*}[ht]
    \centering
    \includegraphics[width=\textwidth]{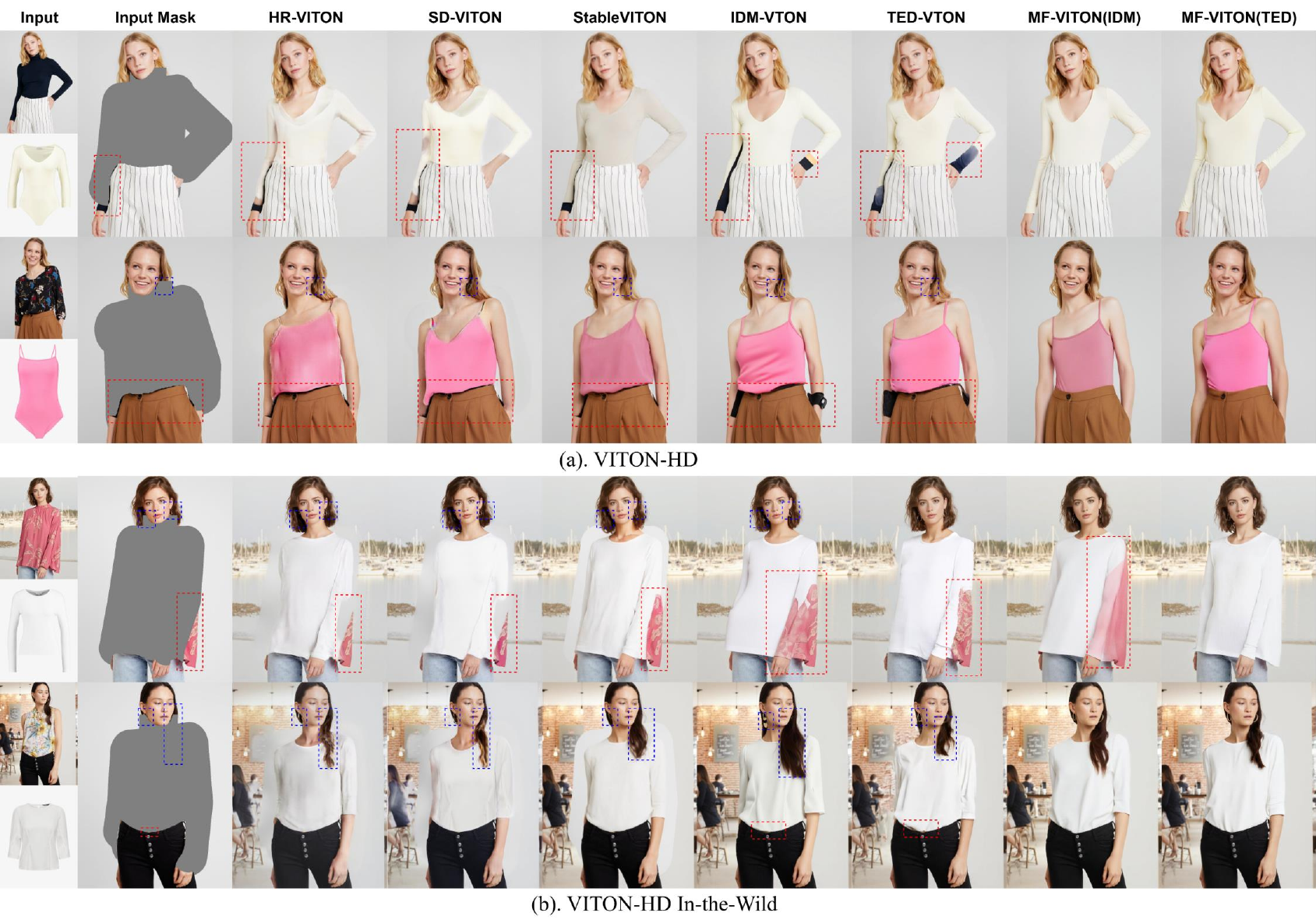}
        \caption{The blue dashed box shows Over-masking, causing person appearance inconsistency, while the red dashed box indicates Mask leakage, introducing artifacts. This figure highlights our model’s superior naturalness and realism compared to SOTA approaches on (a) VITON-HD \cite{choi_viton-hd_2021} and (b) VITON-HD In-the-Wild. All masks are generated and augmented using VITON-HD \cite{choi_viton-hd_2021}. Zoom in for finer details.}
    \label{fig:Qualitative_comparison}
    \vspace{-0.5em}
\end{figure*}

\begin{table*}[ht]
\centering
\vspace{-0.3em}
\resizebox{1\textwidth}{!}{
\begin{tabular}{lccccclccccc}
\hline
\textbf{Dataset} & \multicolumn{5}{c}{\textbf{VITON-HD}} &  & \multicolumn{5}{c}{\textbf{DressCode Upper-body}} \\ \cline{2-6} \cline{8-12} 
\textbf{Method} & LPIPS↓ & SSIM↑ & \multicolumn{1}{l}{CLIP-I↑} & FID↓ UN & KID↓ UN &  & LPIPS↓ & SSIM↑ & \multicolumn{1}{l}{CLIP-I↑} & FID↓ UN & KID↓ UN \\ \hline
\rowcolor[gray]{0.9}\multicolumn{12}{c}{\textbf{Mask-based methods}} \\ \hline
\textbf{HR-VITON \cite{lee_high-resolution_2022}} & 0.115 & 0.877 & 0.800 & 12.238 & 3.757 &  & 0.118 & 0.910 & 0.749 & 29.383 & 3.104 \\
\textbf{SD-VITON \cite{shim_towards_2024}} & 0.104 & \cellcolor{myorange}\textbf{0.896} & 0.831 & 9.857 & 1.450 &  & - & - & - & - & - \\
\textbf{StableVITON \cite{kim_stableviton_2023}} & 0.142 & 0.875 & 0.838 & 9.371 & 1.990 &  & 0.113 & 0.910 & 0.844 & 19.712 & 2.149 \\
\textbf{IDM–VTON \cite{choi_improving_2024}} & 0.102 & 0.868 & 0.875 & 9.156 & 1.242 &  & 0.065 & 0.920 & 0.870 & 11.852 & 1.181 \\
\textbf{TED-VITON \cite{wan_ted-viton_2024}} & \cellcolor{myyellow}{\underline{0.095}} & 0.881 & \cellcolor{myorange}\textbf{0.878} & \cellcolor{myyellow}{\underline{8.848}} & \cellcolor{myyellow}{\underline{0.858}} &  & \cellcolor{myyellow}{\underline{0.050}} & \cellcolor{myyellow}{\underline{0.934}} & \cellcolor{myorange}\textbf{0.875} & 11.451 & 1.393 \\ \hline
\rowcolor[gray]{0.9}\multicolumn{12}{c}{\textbf{Mask-Free methods}} \\ \hline
\textbf{MF-VITON(IDM)} & 0.107 & 0.864 & 0.852 & 9.036 & 1.011 &  & 0.0513 & 0.928 & 0.869 & \cellcolor{myyellow}{\underline{11.361}} & \cellcolor{myorange}\textbf{1.043} \\
\textbf{MF-VITON(TED)} & \cellcolor{myorange}\textbf{0.088} & \cellcolor{myyellow}{\underline{0.886}} & \cellcolor{myyellow}{\underline{0.876}} & \cellcolor{myorange}\textbf{8.441} & \cellcolor{myorange}\textbf{0.560} &  & \cellcolor{myorange}\textbf{0.0412} & \cellcolor{myorange}\textbf{0.939} & \cellcolor{myyellow}{\underline{0.874}} & \cellcolor{myorange}\textbf{11.184} & \cellcolor{myyellow}{\underline{1.107}} \\ \hline
\end{tabular}
}
\caption{Quantitative comparison of models trained and evaluated on VITON-HD and DressCode upper-body datasets. \textbf{Bold} and \underline{underline} denote the best and the second best result, respectively. ``UN'' indicates the unpaired setting, which better reflects real-world usage scenarios where the model must generalize to unseen garment-person combinations. KID score is multiplied by 100.}
\label{tab:hd-result}
\vspace{-1.0em}
\end{table*}

\section{Experiment}

To thoroughly evaluate MF-VITON, we conduct a comprehensive study encompassing both quantitative and qualitative analyses, along with a discussion of key design choices. Our evaluation primarily focuses on (1) the limitations of Mask-based approaches, particularly their susceptibility to mask inaccuracies that introduce unwanted artifacts, and (2) the effectiveness of our Mask-Free strategy in generating high-quality try-on results with accurate garment alignment and fine-grained detail preservation. For quantitative analysis, we employ widely used metrics to assess both the perceptual quality and fidelity of the generated images. These metrics provide a rigorous evaluation of MF-VITON’s ability to produce realistic and visually coherent results while mitigating errors introduced by traditional Mask-based methods. In the qualitative analysis, we compare MF-VITON outputs with baseline models to highlight their superior garment synthesis, background consistency, and robustness in challenging scenarios, such as complex poses and occlusions. In addition, we conduct a discussion to validate the effectiveness of our OFI training strategy, demonstrating its role in improving garment placement accuracy and reducing artifacts. 

\subsection{Experiment Setup}

\noindent\textbf{Baselines}. We evaluated our method in a Mask-Free setting against Mask-based VITON approaches under their respective mask conditions. Mask-based baselines include HR-VITON \cite{lee_high-resolution_2022}, SD-VTON \cite{shim_towards_2024}, StableVITON \cite{kim_stableviton_2023}, IDM-VTON \cite{choi_improving_2024}, and TED-VITON \cite{wan_ted-viton_2024}. These models rely on pre-trained GAN-based generators or SD models with different conditioning techniques. For a fair comparison, we generate images at a resolution of \(1024 \times 768\) when available. Otherwise, we generate images at \(512 \times 384\) and upscale them to \(1024 \times 768\) using interpolation or super-resolution techniques \cite{wang_real-esrgan_2021}, ensuring the highest achievable quality.

\noindent\textbf{Evaluation datasets.} We evaluated the effectiveness of MF-VITON on three datasets: two widely-used VITON benchmarks, VITON-HD \cite{choi_viton-hd_2021} and DressCode \cite{morelli_dress_2022}, and an In-the-Wild dataset created by us, which is an augmented version of VITON-HD. The VITON-HD dataset comprises 13,679 pairs of frontal-view images of women and their corresponding upper garments. Following the standard practices of previous works \cite{morelli_ladi-vton_2023, gou_taming_2023, kim_stableviton_2023, choi_improving_2024, wan_improving_2024, wan_ted-viton_2024}, we split both the original VITON-HD and its In-the-Wild variant into a training set of 11,647 pairs and a test set of 2,032 pairs. The DressCode dataset, which focuses on upper-body garments, contains 15,366 image pairs. In line with its original splits, we use 1,800 upper-body image pairs from DressCode as the test set. All experiments on VITON-HD and DressCode are conducted at a resolution of \(1024 \times 768\), ensuring consistency and comparability with existing studies. 

\noindent\textbf{Evaluation metrics.} We evaluate MF-VITON in both unpaired and paired settings, following prior VITON studies \cite{morelli_ladi-vton_2023, choi_viton-hd_2021, choi_improving_2024, wan_ted-viton_2024}. The unpaired setting, which reflects real-world applications where the input garment differs from the original, is assessed using Fréchet Inception Distance (FID) \cite{heusel_gans_2017} and Kernel Inception Distance (KID) \cite{kim_u-gat-it_2019} to measure realism and distributional similarity. In the paired setting, where the target garment matches the original, we evaluate fidelity using Structural Similarity Index (SSIM) \cite{wang_image_2004} for structural consistency, Learned Perceptual Image Patch Similarity (LPIPS) \cite{zhang_unreasonable_2018} for perceptual similarity, and CLIP image similarity score (CLIP-I) \cite{hessel_clipscore_2021} for semantic alignment.

\subsection{Qualitative Results}

Fig.~\ref{fig:Qualitative_comparison} highlights two common issues in Mask-based virtual try-on approaches: Over-masking (blue dashed boxes), where excessive masking removes critical facial and body details, causing inconsistencies in person appearance, and Mask leakage (red dashed boxes), where incomplete masks allow unwanted artifacts from the original image to persist, reducing realism. These limitations become even more pronounced in real-world scenarios, where background variations further challenge mask accuracy.  

In contrast, the last two columns of Fig.~\ref{fig:Qualitative_comparison} demonstrate the improvements achieved by our MF-VITON models, which eliminate the reliance on segmentation masks. By directly leveraging full-body images as inputs, our approach significantly enhances garment blending, leading to smoother transitions and more natural integration of clothing with the person’s body. Furthermore, our method preserves intricate garment textures and maintains consistency in person identity without introducing visual artifacts.

\begin{table}[t]
    \centering
    \resizebox{1\linewidth}{!}{
        \begin{tabular}{lccccc}
        \hline
        \textbf{Dataset} & \multicolumn{5}{c}{\textbf{VITON-HD In-the-Wild}} \\ \hline
        \textbf{Method} & LPIPS↓ & SSIM↑ & CLIP-I↑ & FID↓ UN & KID↓ UN \\
        \rowcolor[gray]{0.9}\multicolumn{6}{c}{\textbf{Mask-based methods}} \\
        \textbf{HR-VITON \cite{lee_high-resolution_2022}} & 0.183 & 0.850 & 0.823 & 19.593 & 7.214 \\
        \textbf{SD-VITON \cite{shim_towards_2024}} & 0.174 & 0.857 & 0.804 & 17.151 & 5.326 \\
        \textbf{StableVITON \cite{kim_stableviton_2023}} & 0.201 & 0.863 & 0.848 & 12.983 & 2.804 \\
        \textbf{IDM–VTON \cite{choi_improving_2024}} & 0.127 & 0.832 & \cellcolor{myyellow}{\underline{0.931}} & 11.373 & 1.133 \\
        \textbf{TED-VITON \cite{wan_ted-viton_2024}} & 0.113 & 0.861 & 0.927 & 11.239 & \cellcolor{myyellow}{\underline{0.939}} \\
        \rowcolor[gray]{0.9}\multicolumn{6}{c}{\textbf{Mask-Free methods}} \\
        \textbf{MF-VITON(IDM)} & \cellcolor{myyellow}{\underline{0.108}} & \cellcolor{myyellow}{\underline{0.864}} & 0.842 & \cellcolor{myorange}\textbf{9.256} & 1.111 \\
        \textbf{MF-VITON(TED)} & \cellcolor{myorange}\textbf{0.085} & \cellcolor{myorange}\textbf{0.884} & \cellcolor{myorange}\textbf{0.938} & \cellcolor{myyellow}{\underline{10.859}} & \cellcolor{myorange}\textbf{0.918} \\ \hline
        \end{tabular}
    }
    \caption{Quantitative comparison of models on the VITON-HD In-the-Wild dataset. ``UN'' indicated the unpaired setting. KID score is multiplied by 100.}
    \label{tab:result_itw}
    \vspace{-1.5em}
\end{table}

\subsection{Quantitative Results}

Table.~\ref{tab:hd-result} and Table.~\ref{tab:result_itw} illustrate the effectiveness of our MF-enhanced models across multiple benchmarks. On VITON-HD~\cite{choi_viton-hd_2021} and DressCode Upper-body~\cite{morelli_dress_2022}, our approach achieves substantial improvements over non-MF counterparts and existing SOTA methods. Specifically, MF-VITON(IDM) and MF-VITON(TED) consistently exhibit lower perceptual distance (LPIPS), improved realism (FID), and better semantic consistency (KID), highlighting MF’s ability to preserve fine-grained textures and garment fidelity. 

Moreover, in real-world settings, our background-aware training strategy significantly enhances model generalization, as evidenced by the results in Table.~\ref{tab:result_itw}. By reducing perceptual discrepancies and improving semantic alignment, MF-enhanced models achieve seamless garment integration while maintaining background consistency. These consistent performance gains across diverse datasets and environments demonstrate the robustness of our approach in handling both paired and unpaired scenarios.

\subsection{Discussion}

\paragraph{Effectiveness of OFI Strategy}  

\begin{figure}[t]
    \centering
    \includegraphics[width=1\linewidth]{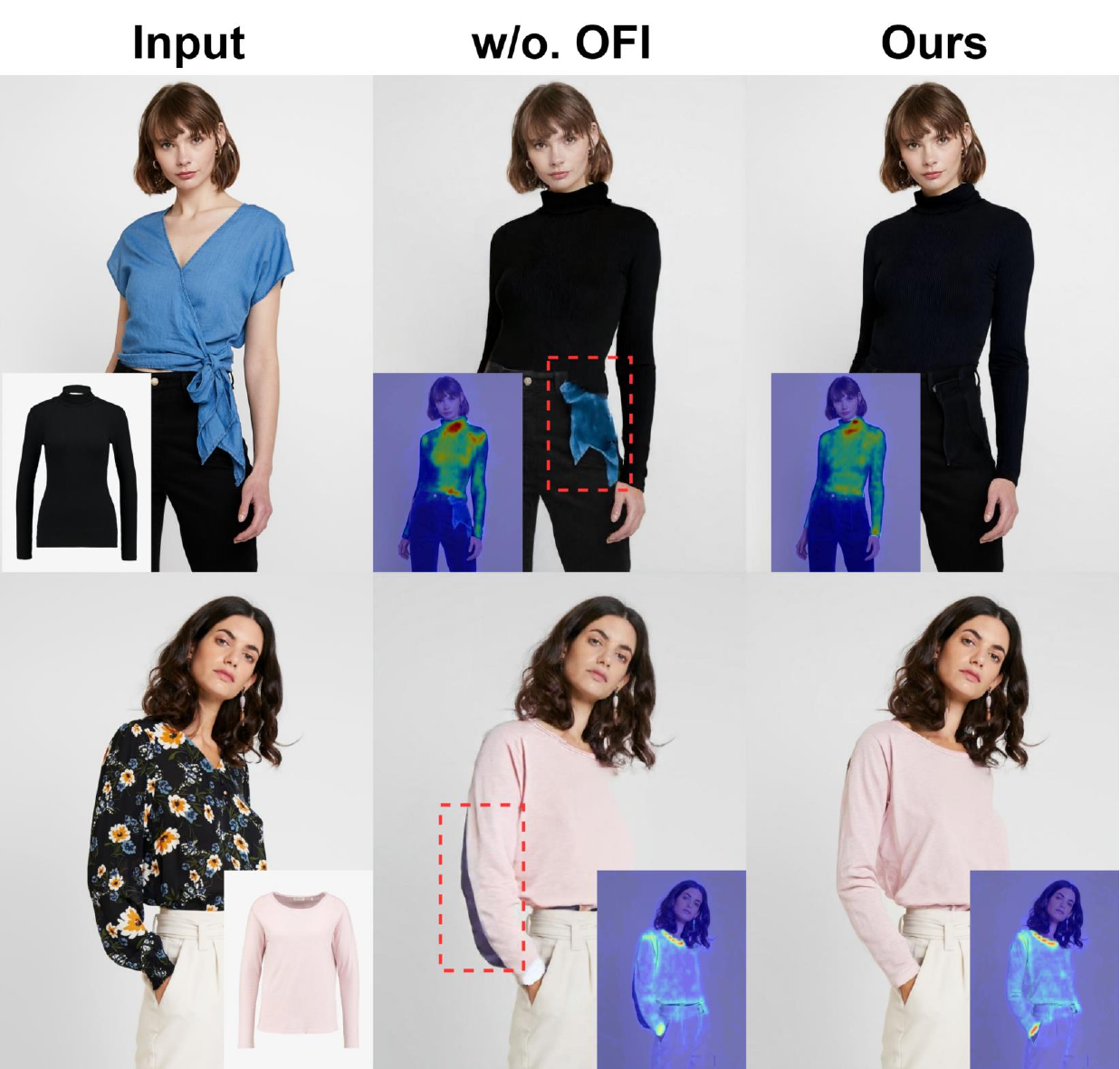}
    \caption{Comparison of virtual try-on results with and without the OFI strategy. The OFI-enhanced model achieves better control over the attention map, leading to more accurate garment placement and fewer artifacts, particularly in challenging regions.}
    \label{fig:OFI}
\vspace{-0.5em}
\end{figure}

To assess the effectiveness of our OFI training strategy, we conduct a quantitative comparison on the VITON-HD dataset. As shown in Table~\ref{tab:result_itw}, our method significantly outperforms the original pairwise training strategy across multiple metrics. Beyond numerical improvements, Fig.~\ref{fig:OFI} visually demonstrates the advantages of OFI, showcasing its ability to mitigate artifacts caused by inaccurate masks, such as inconsistencies in hair, hands, or clothing regions. By treating outputs from Mask-based models as inputs during training, OFI enhances robustness to noisy masks, leading to more realistic garment details and improved visual fidelity.

Additionally, as illustrated by the attention maps in Fig.~\ref{fig:OFI}, the OFI-enhanced model achieves more precise alignment with the clothing regions, improving spatial awareness and garment placement. Importantly, without this strategy, the upper bound of a Mask-Free model is inherently constrained by the quality of the Mask-based model that generates its training data. This limitation is evident in Table~\ref{tab:result_itw}, where models trained without OFI perform significantly worse than TED-VITON \cite{wan_ted-viton_2024} in Table~\ref{tab:hd-result}. These findings confirm that OFI not only improves quantitative performance but also enhances generalization in challenging cases.

\begin{table}[t]
    \centering
    \resizebox{1\linewidth}{!}{
        \begin{tabular}{lccccc}
        \hline
        \textbf{Strategy} & LPIPS↓ & SSIM↑ & CLIP-I↑ & FID↓ UN & KID↓ UN \\ \hline
        \textbf{w/o. OFI} & 0.118 & 0.868 & 0.828 & 12.645 & 4.662 \\
        \textbf{OFI(Ours)} & \textbf{0.088} & \textbf{0.886} & \textbf{0.876} & \textbf{8.441} & \textbf{0.560} \\ \hline
        \end{tabular}
    }
    \caption{Quantitative comparison of models trained on the VITON-HD dataset with different strategies. ``UN'' indicated the unpaired setting. KID score is multiplied by 100.}
    \label{tab:result_itw}
    \vspace{-2em}
\end{table}
\section{Conclusion}
\label{sec:conclusion}

In this work, we introduce MF-VITON, a novel Mask-Free VITON framework that removes the need for mask inputs by leveraging a Mask-Free dataset creation strategy and training process. Our approach overcomes a major limitation of existing VITON methods, which depend heavily on predefined clothing masks. Comprehensive evaluations on VITON-HD \cite{choi_viton-hd_2021}, DressCode \cite{morelli_dress_2022}, and our self-constructed VITON-HD In-the-Wild dataset show that MF-VITON, when combined with the OFI strategy, significantly improves performance while reducing input requirements. Notably, our method achieves superior visual fidelity with fewer input requirements compared to previous Mask-based VITON approaches. These results underscore the efficiency and effectiveness of MF-VITON, establishing it as a practical and scalable solution for real-world VITON applications.

{
\small
\bibliographystyle{ieeenat_fullname}

}


\end{document}